\documentclass[runningheads]{llncs}
\usepackage{graphicx}
\usepackage{blindtext}
\usepackage[font=small,labelfont=bf]{caption}
\usepackage{multirow}
\usepackage{amssymb}
\usepackage{xcolor}
\usepackage{esvect}
\usepackage{float}
\usepackage{subfig}
\usepackage[tableposition=top]{caption}
\begin{document}
\title{Invertible Network for Classification and Biomarker Selection for ASD}
%
%
\author{ Juntang Zhuang$^1$, \quad Nicha C. Dvornek $^{3}$,\quad  Xiaoxiao Li$^1$, \quad Pamela Ventola$^2$,\qquad James S. Duncan$^{1,3,4}$}
\footnotesize{
 \institute{$^{1}$ Biomedical Engineering, Yale University, New Haven, CT USA  \\ $^{2}$ Child Study Center, Yale University, New Haven, CT USA\\ $^{3}$ Radiology \& Biomedical Imaging, Yale School of Medicine, New Haven, CT USA \\  $^{4}$ Electrical Engineering, Yale University, New Haven, CT USA}
}
%
%
\maketitle              
\begin{abstract}
Determining biomarkers for autism spectrum disorder (ASD) is crucial to understanding its mechanisms. Recently deep learning methods have achieved success in the classification task of ASD using fMRI data. However, due to the black-box nature of most deep learning models, it's hard to perform biomarker selection and interpret model decisions. The recently proposed invertible networks can accurately reconstruct the input from its output, and have the potential to unravel the black-box representation. Therefore, we propose a novel method to classify ASD and identify biomarkers for ASD using the connectivity matrix calculated from fMRI as the input. Specifically, with invertible networks, we explicitly determine the decision boundary and the projection of data points onto the boundary. Like linear classifiers, the difference between a point and its projection onto the decision boundary can be viewed as the \textit{explanation}. We then define the \textit{importance} as the explanation weighted by the gradient of prediction $w.r.t$ the input, and identify biomarkers based on this importance measure. We perform a regression task to further validate our biomarker selection: compared to using all edges in the connectivity matrix, using the top 10\% important edges we generate a lower regression error on 6 different severity scores. Our experiments show that the invertible network is both effective at ASD classification and interpretable, allowing for discovery of reliable biomarkers.

\keywords{invertible network, ASD, biomarker, regression}
\end{abstract}

\section{Introduction}
Autism spectrum disorder (ASD) is a neurodevelopmental disorder that affects social interaction and communication, yet the causes for ASD are still unknown \cite{speaks2011autism}. Functional MRI (fMRI) can measure the contrast dependent on blood oxygenation \cite{ogawa1990brain} and reflect brain activities, and therefore has the potential to help in understanding ASD. 

Recent research efforts have applied machine learning and deep learning methods on fMRI data to classify ASD versus control groups \cite{anderson2011functional,heinsfeld2018identification} and predict treatment outcomes \cite{zhuang2018prediction,zhuang2018prediction2}. However, deep learning models are typically hard to interpret, and thus are difficult to use for identifying biomarkers for ASD. 

Various methods have been proposed to interpret a deep neural network. Bach et al. proposed to assign the decision of a neural network to its input with layer-wise relevance propagation \cite{bach2015pixel}; Mahendran et al. proposed to approximately invert the network to explain its decision \cite{mahendran2015understanding}; Sundararajan et al. proposed integrated gradient to explain a model's decision \cite{sundararajan2017axiomatic}. However, all these methods only generate an approximation to the inversion of neural networks, and can not unravel the black-box representation of deep learning models. 


The recently proposed invertible network can accurately reconstruct the input from its output \cite{jacobsen2018revnet,behrmann2018invertible}. Based on this property, we propose a novel method to  interpret ASD  classification on fMRI data. As shown in Fig.~\ref{fig:ivnnet}, an invertible network  first maps data from the \textit{input} domain (e.g. connectivity matrix calculated from fMRI data) to the \textit{feature} domain, then applies a fully-connected layer to classify ASD from control group. Since a fully-connected layer is equivalent to a linear classifier in the feature domain, we can determine the decision boundary as a high-dimensional plane, and calculate projection of a point onto the boundary. Since our network is  invertible, we can invert the decision boundary in the \textit{feature} domain to the \textit{input} domain. As shown in Fig.~\ref{fig:projection}, the difference between the input and its projection on the decision boundary can be viewed as the explanation for the model's decision. We applied the proposed method on ASD classification of the ABIDE dataset achieving 71\% accuracy, and then identified biomarkers for ASD based on the proposed interpretation method. We further validated the selected biomarkers (edges in the connectivity matrix as in Fig.~\ref{fig:connectome} and ROIs as in Fig.~\ref{fig:roi})  in a regression task: compared with using all edges, the selected edges generate more accurate predictions for ASD severity (Table~\ref{table:regression}).


Our contributions can be summarized as:
\begin{itemize}
    \item Based on the invertible network, we proposed a novel method to intepret model decision and rank feature importance. 
    \item We applied the proposed method on an ASD classification task, achieved a high classification accuracy (71\% on the whole ABIDE dataset), and identified biomarkers.
    \item We demonstrated effectiveness of selected biomarkers in a regression task.
\end{itemize}
\section{Methods}
\begin{figure}[]
  \centering
  \begin{minipage}[b]{0.7\textwidth}
    \includegraphics[width=\textwidth]{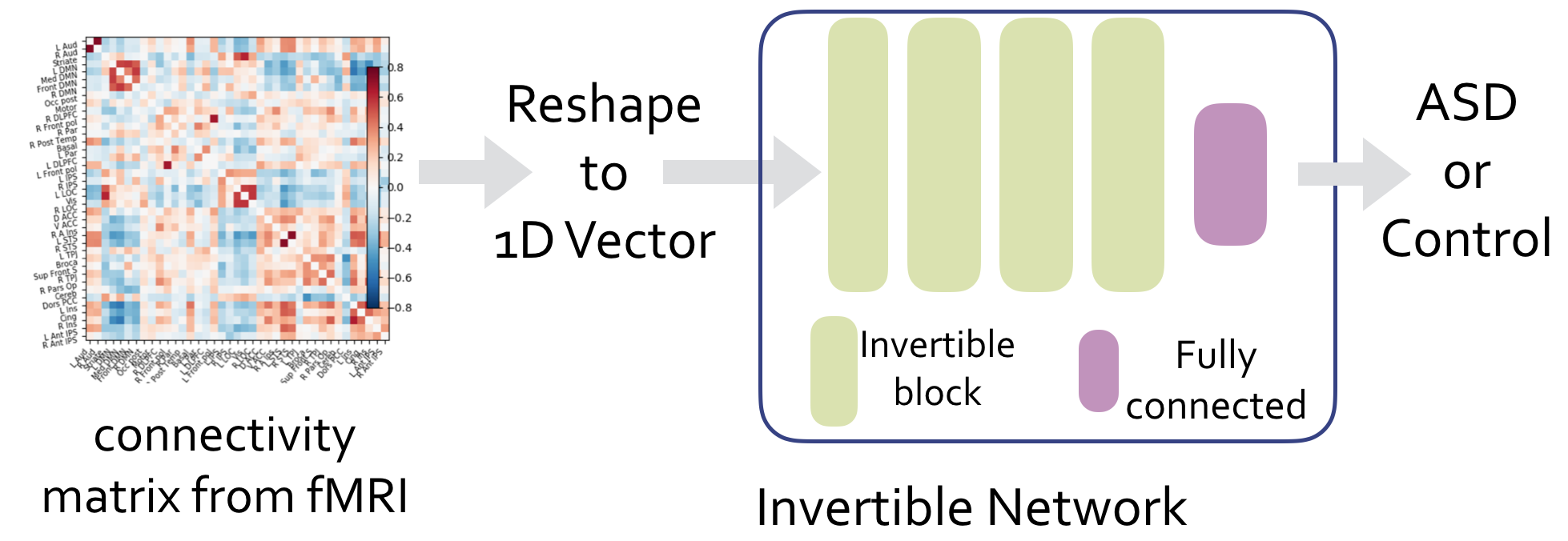}
    \caption{\small{Classification pipeline for fMRI and structure of invertible network.}}
    \label{fig:ivnnet}
  \end{minipage}
  \hfill
  \begin{minipage}[b]{0.28\textwidth}
    \includegraphics[width=\textwidth,height=2.3cm]{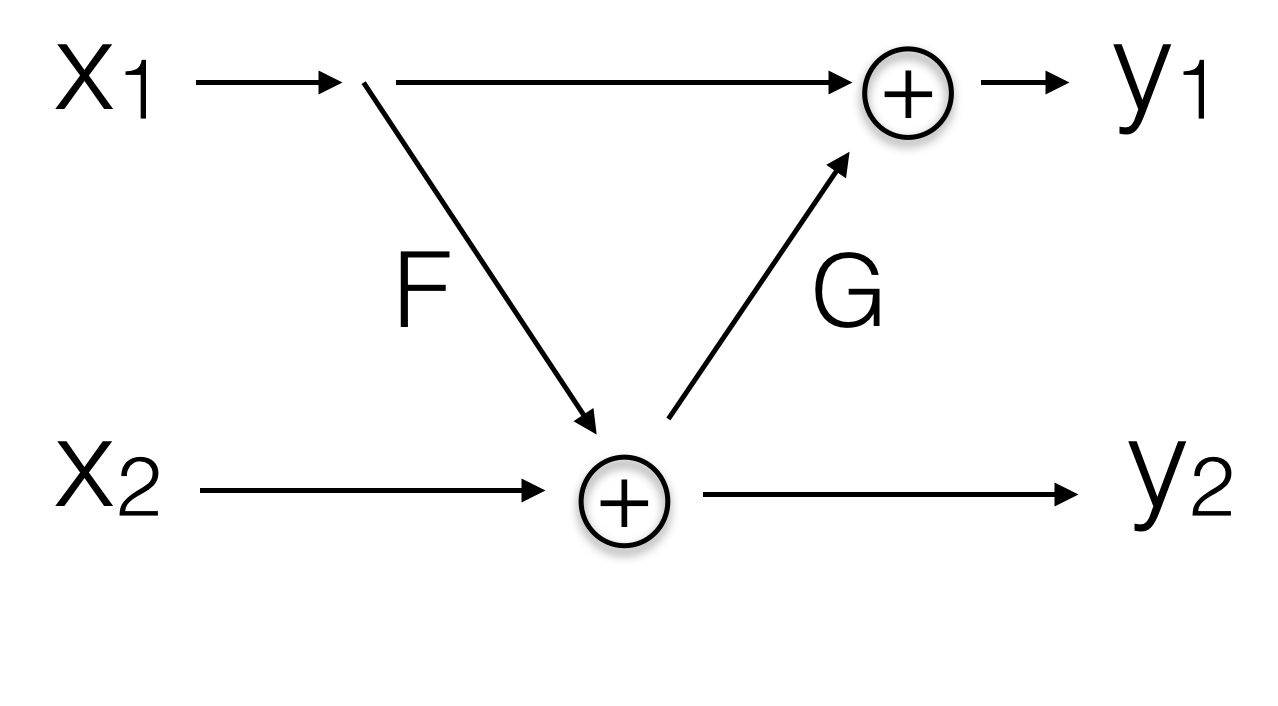}
    \caption{\small{Structure of invertible blocks.}}
    \label{fig:inv_block}
  \end{minipage}
\end{figure}

The classification pipeline is summarized in Fig.~\ref{fig:ivnnet}. In this section, we first discuss input data and pre-processing of fMRI, then introduce the structure of invertible networks as in Fig.~\ref{fig:ivnnet}. Next, we propose a novel method to determine the decision boundary and identify biomarkers for ASD. 
\subsection{Dataset and Inputs}
The ABIDE dataset \cite{di2014autism} consists of fMRI data for 530 subjects with ASD and 505 control subjects. We use the pre-processed data by the Connectome Computation System (CCS) \cite{zuo2013toward} pipeline, which includes registration, 4D global mean-based intensity normalization, nuisance correction and band-pass filtering. We extract the mean time series for each region of interest (ROI) defined by the CC200 atlas \cite{craddock2012whole}, consisting of 200 ROIs. We compute the Pearson Correlation between the time series of every pair of ROIs, and reshape the connectivity into a vector of length $200 \times 199/2 = 19900$. This vector is the input to the invertible network.
\subsection{Invertible Networks}
\label{sec:InvNet}
The structure of the invertible network is shown in Fig.~\ref{fig:ivnnet}. An invertible network is composed of a stack of invertible blocks and a final fully-connected (FC) layer to perform classification. Note that inversion does not update parameters, thus is different from the backward propagation method; invertible networks can be trained with the backward propagation method as normal networks.
\subsubsection{Invertible Blocks} 
An invertible block is shown in Fig.~\ref{fig:inv_block}, where the input is split by channel into two parts $x_1$ and $x_2$, and the outputs are denoted as $y_1$ and $y_2$. Feature maps $x_1,x_2,y_1,y_2$ have the same shape. $F$ and $G$ are \textit{non-linear} functions with parameters to learn: for 1D input, $F$ and $G$ can be a sequence of FC layers, 1D batch normalization layers and activation layers; for 2D input, $F$ and $G$ can be a sequence of convolutional layers, 2D batch normalization layers and activation layers. $F$ and $G$ are required to generate outputs of the same shape as input. The invertible block can accurately recover the input from its output, where the forward pass and inverse of an invertible block are:
\begin{equation} \label{eq_rev_forward}
   \left\{
                \begin{array}{ll}
                  y_2 = x_2 + F(x_1)\\
                  y_1 = x_1 + G(y_2)\\
                \end{array}
    \right.
    \left\{
                \begin{array}{ll}
                  x_1 = y_1 - G(y_2)\\
                  x_2 = y_2 - F(x_1)\\
                \end{array}
    \right.
\end{equation}
\subsubsection{Notations}
The invertible network classifier can be viewed as a 2-stage model:
\begin{equation}\label{eq:notation}
    z = T(x), y = C(z)
\end{equation}
where $T$ is the invertible transform and $C$ is the final FC layer which is a $linear$ classifier. We denote $x\in \mathbb{R}^d$ as the input domain (e.g. connectivity matrix from fMRI reshaped to 1D vector), and $z\in \mathbb{R}^d$ as the feature domain. A data point is mapped from the input domain to the feature domain by an \textit{invertible} transform $T$, then a \textit{linear} classifer $C$ is applied.
\subsection{Model Decision Interpretation and Biomarker Selection}
In this section we propose a novel method to interpret decisions of an invertible network classifier. We begin with a linear classifier, then generalize to non-linear classifier such as neural networks. An example is shown in Fig.~\ref{fig:projection}.
\subsubsection{Interpret Decision of a Linear Classifier}
A linear classifier is calculated as:
\begin{equation} \label{eq:linear_classifier}
    f(x)=\left\{
                \begin{array}{ll}
                  1\ \ \ if\ \langle \vec{w}, x \rangle + b >0\\
                  0\ \ \ if\ \langle \vec{w}, x \rangle + b \leq 0
                \end{array}
              \right.
\end{equation}
where $\vec{w}$ is the weight vector, and $b$ is the bias.
The decision boundary is a high dimensional plane, and can be denoted as 
$ 
    \{x: \langle \vec{w},x \rangle +b =0\}
$. 

For a data point $x$, we calculate its projection onto the decision boundary as
\begin{equation} \label{eq:projection}
x_p = x - \langle \frac{\vec{w}}{\vert \vert \vec{w} \vert \vert_2}, x\rangle \frac{\vec{w}}{\vert \vert \vec{w} \vert \vert_2} - b \frac{\vec{w}}{\vert \vert \vec{w} \vert \vert_2^2} \hfill
\end{equation}
and define the $explanation$ and $importance$ as:
\begin{equation}\label{eq:explanation}
    explanation = x - x_p,\ \
importance = \vert \vec{w} \otimes (x - x_p) \vert
\end{equation}
where $\otimes$ is element-wise product, the difference $x-x_p$ can be viewed as the explanation for the linear classifier, and the $importance$ of each input dimension is defined as the absolute value of the $explanation$ weighted by $\vec{w}$. An example is shown in Fig.~\ref{fig:projection}(a).
\subsubsection{Determine Projection onto Decision Boundary of Invertible Networks}
As in Equation \ref{eq:notation}, a point $x$ in the input domain is mapped to the feature domain, denoted as $X=T(x)$. Since the classifier is linear in the feature domain, we can calculate the projection of $X$ onto the boundary as in Equation \ref{eq:projection}, denoted as $X_p$. Note that since $T$ is invertible, we can map it to the input domain, denoted as $x_p = T^{-1}(X_p)$ where $T^{-1}$ is the inverse operation of $T$ as in Equation \ref{eq_rev_forward}. Furthermore, we can invert the \textit{decision boundary} from the feature domain to the input domain.

\begin{figure}[h]
\centering
\includegraphics[width=\linewidth]{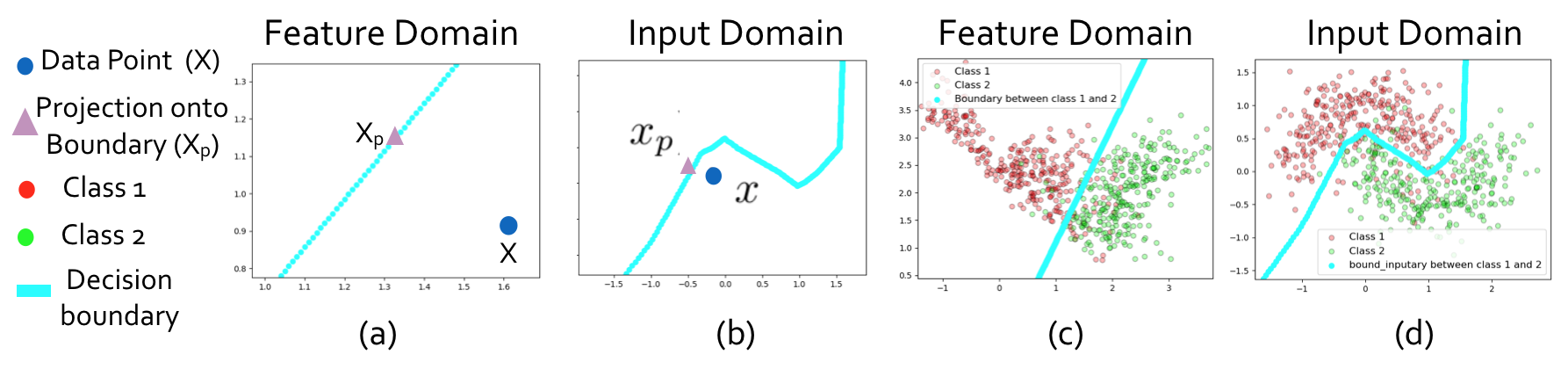}
\caption{
\small{Illustration on a simulation dataset. Fig. (a): data point $X$ and its projection $X_p$ onto the decision boundary. In the feature domain the model is a linear classifier. Fig. (b) Corresponding points $x$ and $x_p$ in the input domain, calculated as inversion of points from (a). Fig. (c)  and (d): Points from two classes are sampled around two interleaving half circles. Decision boundary is a line in the feature domain (c), and is a curve in the input domain (d).}}
\label{fig:projection}
\end{figure}

An example is shown in Fig.~\ref{fig:projection}. Fig.~\ref{fig:projection}(a) shows a data point $X$ and its projection $X_p$ in the feature domain, Fig.~\ref{fig:projection}(b) inverts the projection back to the input domain. Fig.~\ref{fig:projection}(c) and (d) show the decision boundary in the feature and input domain, respectively.
\subsubsection{Feature Importance and Biomarker Selection}
We expand the network $f$ around current point $x$ using Taylor expansion:
\begin{equation} \label{eq:taylor_expansion}
    f(x+\Delta x) = f(x) + \langle \nabla f(x) ,\Delta x \rangle + o(\Delta x)
\end{equation}
Similar to Equation \ref{eq:explanation}, we approximate $f$ locally as a linear classifier and define the $explanation$ and $importance$ as:
\begin{equation}
\label{eq:explain}
explanation = x - x_p,\ \ importance = \vert \nabla f(x) \otimes (x - x_p) \vert
\end{equation}

Equation \ref{eq:explain} defines individualized importance for each data point, and we calculate the mean importance across the entire dataset. We select edges with top importance values as the biomarkers for ASD. 

\begin{figure}[H]
\centering
\includegraphics[width=1\linewidth]{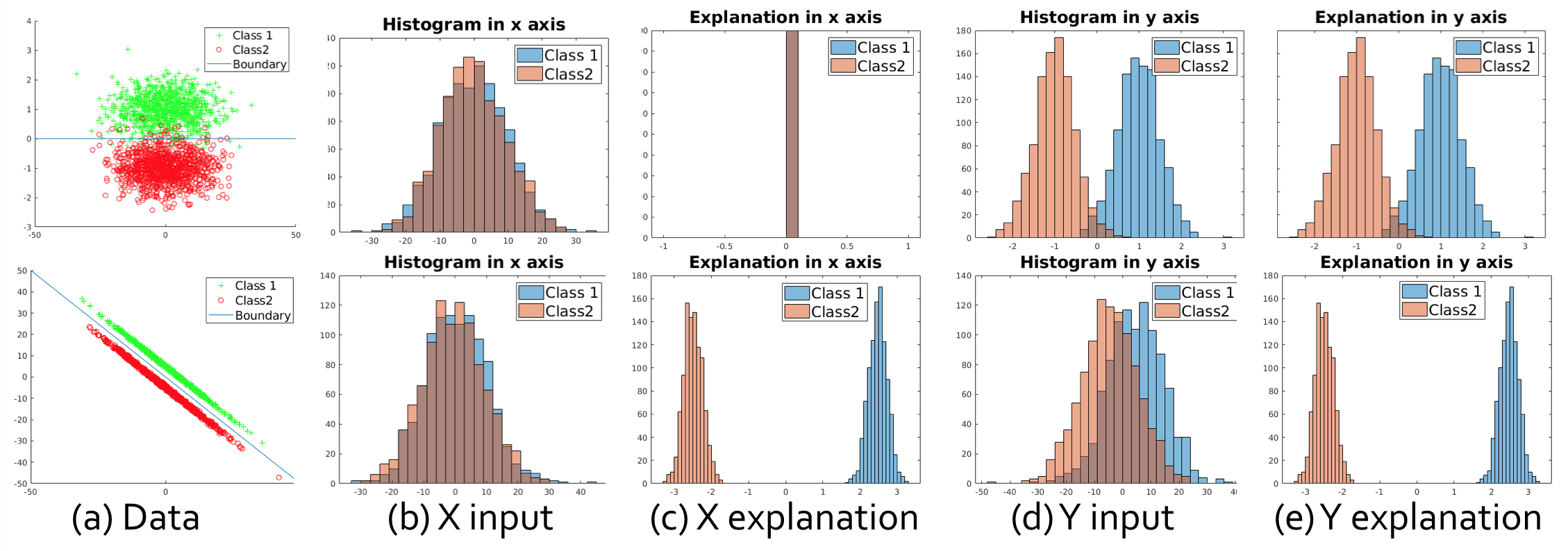}
\caption{\small{
Explanation method on simulation datasets. Columns (a) to (e) represent data distribution, input value in $x$ axis, explanation in $x$ axis, input value in $y$ axis and explanation in $y$ axis. In the top row, both input and explanation fail to distinguish two classes in $x$ axis, but succeed in $y$ axis. In the bottom row, both $x$ and $y$ values are useful; two clusters are \textit{overlapped} in distribution of $x$ ($y$) \textit{input} values, but \textit{separated} in the \textit{explanation} for $x$ ($y$) axis.}}
\label{fig:explain}
\end{figure}

Note that two classes may be close in the input but separated in the explanation. We show two examples in Fig.~
\ref{fig:explain}. In the top row, the decision boundary is a horizontal line, thus $y$ is useful to distinguish the two clusters while $x$ is not. In this case, the distribution of the two clusters are overlapped in the $x$ axis but separated in the $y$ axis for both input and the explanation. In the bottom row, two clusters are separated by line $y=-x$: neither $x$ nor $y$ axis can distinguish two clusters using the input (Fig.~\ref{fig:explain}(b),(d)), but both axes have a large separation margin using the explanation(Fig.~\ref{fig:explain}(c),(e)).

\section{Experiments}
\subsubsection{Classification Accuracy}
We perform a 10-fold cross validation on the entire ABIDE dataset to classify ASD from control group. We compare the invertible network classifier with other methods including SVM, random forest (RF) with 5,000 trees, a 1-layer multi-layer perceptron (MLP) and a 2-layer MLP. Our invertible network (InvNet) has 2 invertible blocks, and the $F$ and $G$ in an invertible block is FC-ReLU-FC-ReLU, where the first FC layer maps a 19,900 (edges in a connectivity matrix using CC200 atlas with 200 ROIs \cite{craddock2012whole}) dimensional vector to a 512 dimensional vector, and the second FC maps a vector from 512 dimension to 19,900. For each of the 1035 subjects, we augment data 50 times by bootstrapping voxels within each ROI, then calculating the connectome from the bootstrapped mean time series. This results in 51570 examples. All deep learning models are trained with SGD optimizer for 50 epochs, with a learning rate of 1e-5 and a cross-entropy loss. 

Numerical results are summarized in Table~\ref{table:accuracy}. Compared with other methods including a Deep Neural Network (DNN) model in \cite{heinsfeld2018identification}, the proposed InvNet generates the highest accuracy (0.71), recall (0.71) and F1 score (0.71). 
\begin{table}
\parbox{.38\linewidth}{
\centering
\captionof{table}{\small{
      Classification results}}
      \label{table:accuracy}
\scalebox{0.65}{
    \begin{tabular}{l|llllll}
\hline
          & SVM  & RF   & MLP(1) & MLP(2) & DNN  &  InvNet \\ \hline
Accuracy  & 0.67 & 0.66 & 0.66       & 0.68        & 0.70 & \textbf{0.71}  \\
Precision & 0.68 & 0.65 & 0.66       & 0.68        & 0.74 & \textbf{0.72}  \\
Recall    & 0.68 & 0.65 & 0.66       & 0.67        & 0.63 & \textbf{0.71}  \\
F1        & 0.68 & 0.68 & 0.67       & 0.69        & -    & \textbf{0.71}  \\ \hline
\end{tabular}
}
}
\hfill
\parbox{.6\linewidth}{
\centering
\caption{\small{Regression results for ASD severity scores}}
\scalebox{0.6}{
\begin{tabular}{l|l|llllll}
\hline
                     & \# Edges & Total & Awareness & Cognition & Comm  & Motivation & Mannerism \\ \hline
\multirow{3}{*}{MSE} & 100\%    & 44.2  & 3.86      & 6.90      & 11.95 & 7.59       & 6.37      \\
& RF(top 10\%) & 71.8 & 8.2 & 12.6 & 22.1 & 12.7 & 11.8 \\
                     & \textbf{ours}     & \textbf{42.8}  & \textbf{3.72}      & \textbf{6.84}      & \textbf{11.70} & \textbf{7.17}       & \textbf{6.19}      \\ \hline
\multirow{3}{*}{Cor} & 100\%    & 0.14  & 0.33      & 0.38      & 0.20  & -0.03      & 0.22      \\
& RF(top 10\%) & 0.09 & -0.19 & 0.01 & -0.01 & -0.12 & -0.08 \\
                     & \textbf{ours}     & \textbf{0.18}  & \textbf{0.42}      & \textbf{0.40}      & \textbf{0.28}  & \textbf{0.20}       & \textbf{0.32}      \\ \hline
\end{tabular}
}
\label{table:regression}
}
\end{table}
\begin{figure*}
\centering
\includegraphics[width=0.9\linewidth]{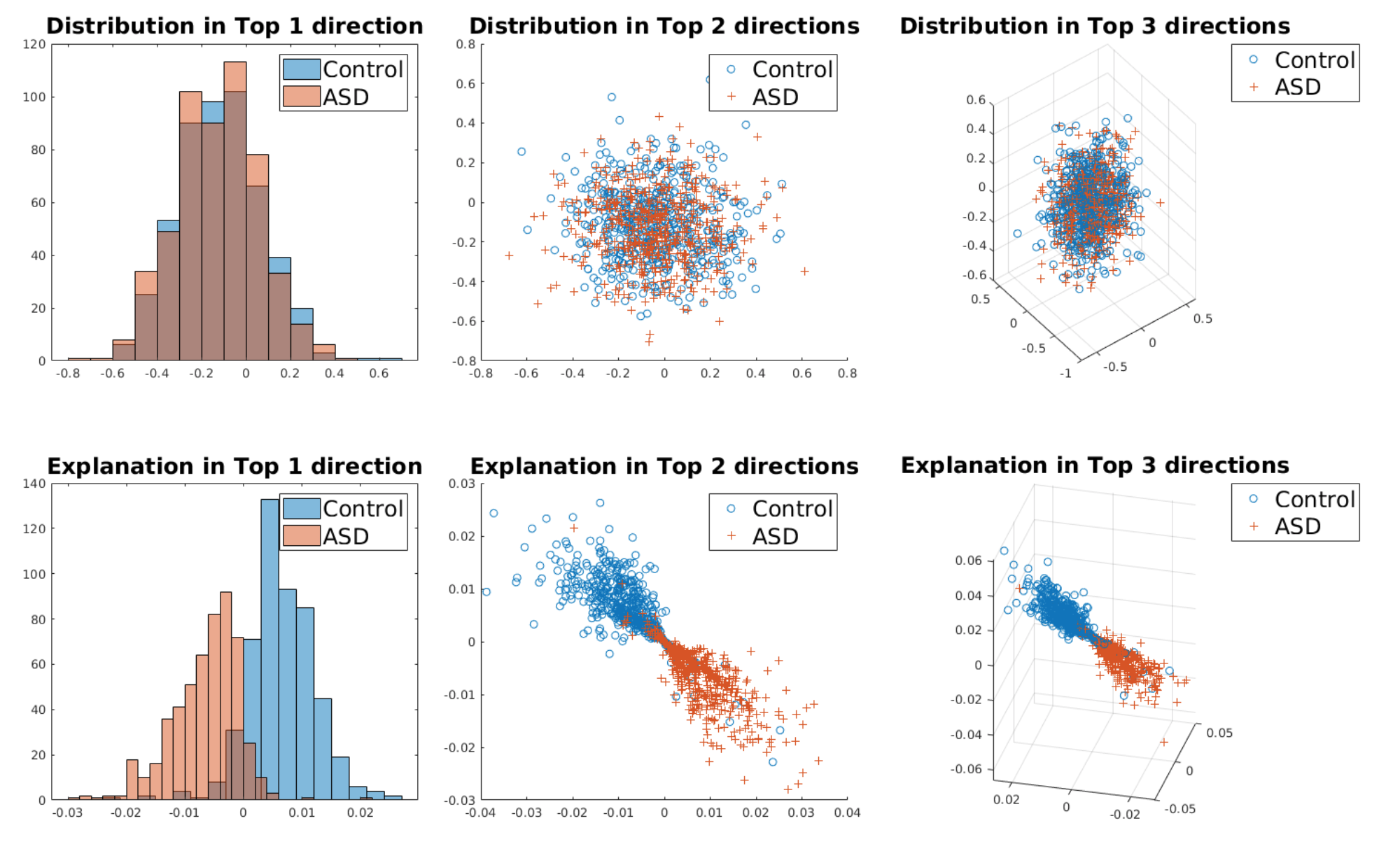}
\caption{\small{ Top row: distribution of connectivity edge for two groups (ASD v.s. Control). Bottom row: distribution of explanation for two groups. From left to right: results on edges with top 1, top 2 and top 3 importance defined in Equation 
\ref{eq:explain}.
\label{fig:explain_ASD}
}}
\end{figure*}
\subsubsection{Explanation and Biomarker Selection}
We plot the histogram of explanation for edges with top 3 importance in Fig.~\ref{fig:explain_ASD}. The histogram of edge values (top row) can not  distinguish ASD from control group, while the distribution of explanations (bottom row) for the two groups are separated. The proposed explanation method can be viewed as a naive embedding method, mapping data to a low-dimensional space where two classes are separated.

For each edge, we calculate the importance as defined in Equation \ref{eq:explain} and select the top 20 connections, as shown in Fig.~\ref{fig:connectome}, and plot the corresponding ROIs in Fig.~\ref{fig:roi}. The proposed method found many regions that are shown to be closely related to autism in the literature, including: superior temporal gyrus \cite{bigler2007superior}, frontal cortex \cite{courchesne2005frontal,carper2005localized}, precentral gyrus \cite{nebel2014precentral}, insular cortex \cite{gogolla2014sensory} and other regions. The selected edges and ROIs can be viewed as biomarkers for ASD. Detailed results are in the appendix.
\newline
\newline
\begin{minipage}{.48\textwidth}
\centering
  \includegraphics[width=0.85\linewidth]{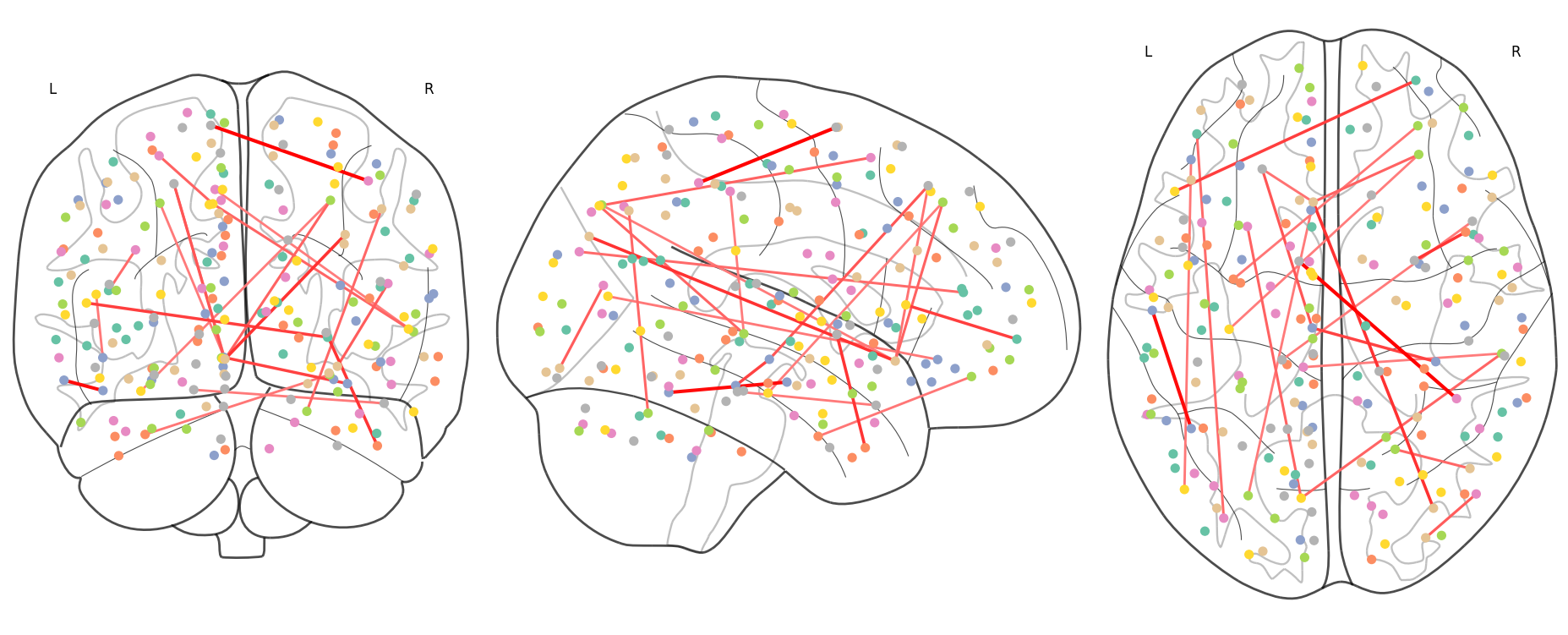}
  \captionof{figure}{\small{Top 20  connections selected by the proposed method.}}
  \label{fig:connectome}
\end{minipage}%
\hfill
\begin{minipage}{.48\textwidth}
  \centering
  \includegraphics[width=\linewidth]{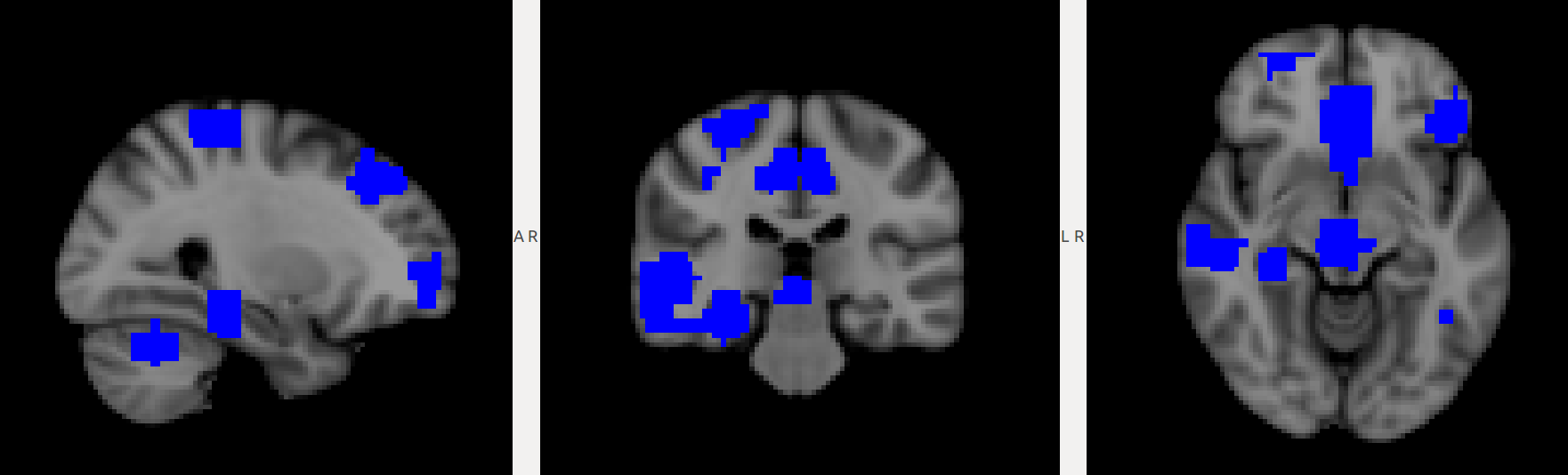}
  \captionof{figure}{\small{Top ROIs selected by the proposed method.}}
  \label{fig:roi}
\end{minipage}
\subsubsection{Validate Biomarker Selection in Regression}
We validate our biomarker selection in a regression task on the ABIDE dataset. The input is the connectivity matrix reshaped to a vector, and the output is different subscores of the social responsiveness scale (SRS) \cite{constantino2013social}. SRS provides a continuous measure of different aspects of social ability, including a total score and subscores for awareness, cognition, communication, motivation and mannerism. We use the top 10\% of edges selected by the proposed method, and compared its performance with using 100\% of edges. We perform a 10-fold cross validation with linear support vector regression (SVR) using $l_2$ penalty, and within each fold the penalty parameter is chosen by nested cross-validation (choices for $l_2$ penalty strength are $0,10^{-6},10^{-5},...10^{-1}$). 

We validate our model by comparison with other feature selection methods.  We first selected top 10\% important features with RF based on out-of-bag prediction importance with default parameters in MATLAB, then refitted a RF using selected features. Results are marked as RF(top 10\%) in Table 2.

We calculate the mean squared error (MSE) and cross correlation (Cor) between predictions and measurements. Results are summarized in Table \ref{table:regression}. Compared with other methods, our selected biomarkers consistently generates a smaller MSE and a higher Cor, validating our biomarker selection method.
\subsubsection{Generalization to Convolutional Invertible Networks}
We generalize the proposed model decision interpretation method to convolutional networks, and validate it on an image classification task with the MNIST dataset. Results are summarized in the appendix. The proposed method generates more intuitive explanation results on 2D images.
\section{Conclusions}
We introduced a novel decision interpretation and feature importance ranking method based on invertible networks. We then applied the proposed method on a classification task to classify ASD from control group based on fMRI scans. We selected important connections and ROIs as biomarkers for ASD, and validated these biomarkers in a regression task on the ABIDE dataset. Our invertible network generates a high classification accuracy, and our biomarkers consistently generate a smaller MSE and a higher Cor compared with using all edges for regression tasks on different severity measures. The proposed interpretation method is generic, and has the potential in other aspects of interpretable deep learning such as 2D image classification.
\subsubsection*{Acknowledgement}
This research
was funded by the National Institutes of Health (NINDS-R01NS035193).
%
%
%
\bibliographystyle{splncs04}
\bibliography{refs}
\end{document}